%% file: ms.tex
\documentclass[acmtog,authorversion,nonacm]{acmart}

\usepackage{booktabs} 
\usepackage[normalem]{ulem}

\usepackage{graphicx} 

\usepackage[ruled]{algorithm2e} 

\SetAlFnt{\small}
\SetAlCapFnt{\small}
\SetAlCapNameFnt{\small}
\SetAlCapHSkip{0pt}

\acmJournal{TOG}

\setcopyright{acmlicensed}\acmConference[SA Conference Papers '23]{SIGGRAPH Asia 2023 Conference Papers}{December 12--15, 2023}{Sydney, NSW, Australia}
\acmDOI{10.1145/3610548.3618212}

\definecolor{agreen}{rgb}{.2,.65,.2}

\usepackage{graphicx}
\usepackage{amsmath}
\usepackage{booktabs}
\usepackage{xcolor}

\begin{document}

\title{VET: Visual Error Tomography for Point Cloud Completion and High-Quality Neural Rendering}

\author{Linus Franke}
\email{linus.franke@fau.de}
\orcid{0000-0001-8180-0963}
\affiliation{%
  \institution{Friedrich-Alexander-Universit\"at Erlangen-N\"urnberg}
  \country{Germany}
}

\author{Darius Rückert}
\email{darius.rueckert@fau.de}
\orcid{0000-0001-8593-3974}
\affiliation{%
  \institution{Friedrich-Alexander-Universit\"at Erlangen-N\"urnberg}
  \country{Germany}
}
\affiliation{%
  \institution{Voxray GmbH}
  \country{Germany}
}

\author{Laura Fink}
\email{laura.fink@fau.de}
\orcid{0009-0007-8950-1790}
\affiliation{%
  \institution{Friedrich-Alexander-Universit\"at Erlangen-N\"urnberg}
  \country{Germany}
}
\affiliation{%
  \institution{Fraunhofer IIS}
  \country{Germany}
}

\author{Matthias Innmann}
\email{matthias.innmann@navvis.com}
\orcid{0009-0000-5314-1200}
\affiliation{%
  \institution{NavVis GmbH}
  \country{Germany}
}

\author{Marc Stamminger}
\email{marc.stamminger@fau.de}
\orcid{0000-0001-8699-3442}
\affiliation{%
  \institution{Friedrich-Alexander-Universit\"at Erlangen-N\"urnberg}
  \country{Germany}
}

\begin{teaserfigure}
\centering
\includegraphics[width=\linewidth]{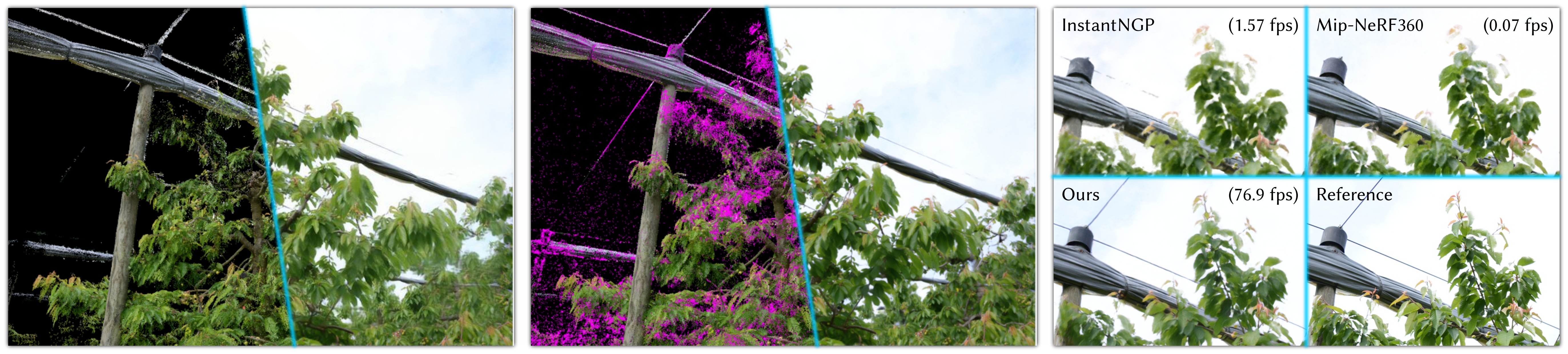}
\caption{The initial point cloud of the \textsc{Cherry Tree} lacks completeness, with branches and leaves missing. Neural point-based rendering exhibits severe artifacts in these regions \textit{(left)}. %
By applying VET and spawning new points, we can complete the point cloud efficiently. 
This appearance-driven geometric completion improves quality of the neural rendering significantly \textit{(middle)} thus outperforming state of the art methods \textit{(right)}.\vspace{3mm}
}%
\label{fig:teaser}
\end{teaserfigure}

\input{00-abstract.tex}

%
%


%
%


\maketitle
\begin{small}
This is the author's version of the work. 
It is posted here for your personal use. 
Not for redistribution. 
The definitive Version of Record was published in SiggraphAsia'23, \url{http://dx.doi.org/10.1145/3610548.3618212}

\hspace{5cm}
\end{small}

\input{01-intro.tex}
\input{02-related.tex}

\input{03-main.tex}
\input{04-eval.tex}

\input{05-conc.tex}

\begin{acks}
The authors thank the anonymous reviewers for their valuable feedback and Lukas Meyer for the \textsc{Cherry Tree} scene.
Additionally, we thank Stefan Romberg, Michael Gerstmayr and Tim Habigt for the fruitful discussions as well as NavVis GmbH for providing datasets for testing.

Linus Franke was supported by the Bayerische Forschungsstiftung (Bavarian Research Foundation) AZ-1422-20.
The authors gratefully acknowledge the scientific support and HPC resources provided by the Erlangen National High Performance Computing Center (NHR@FAU) of the Friedrich-Alexander-Universit\"at Erlangen-N\"urnberg (FAU) under the NHR project \textit{b162dc}. NHR funding is provided by federal and Bavarian state authorities. NHR@FAU hardware is partially funded by the German Research Foundation (DFG) – 440719683.
\end{acks}

\bibliographystyle{ACM-Reference-Format}
\bibliography{egbib}

\newpage

\begin{figure*}
	\centering
	\includegraphics[width=\linewidth]{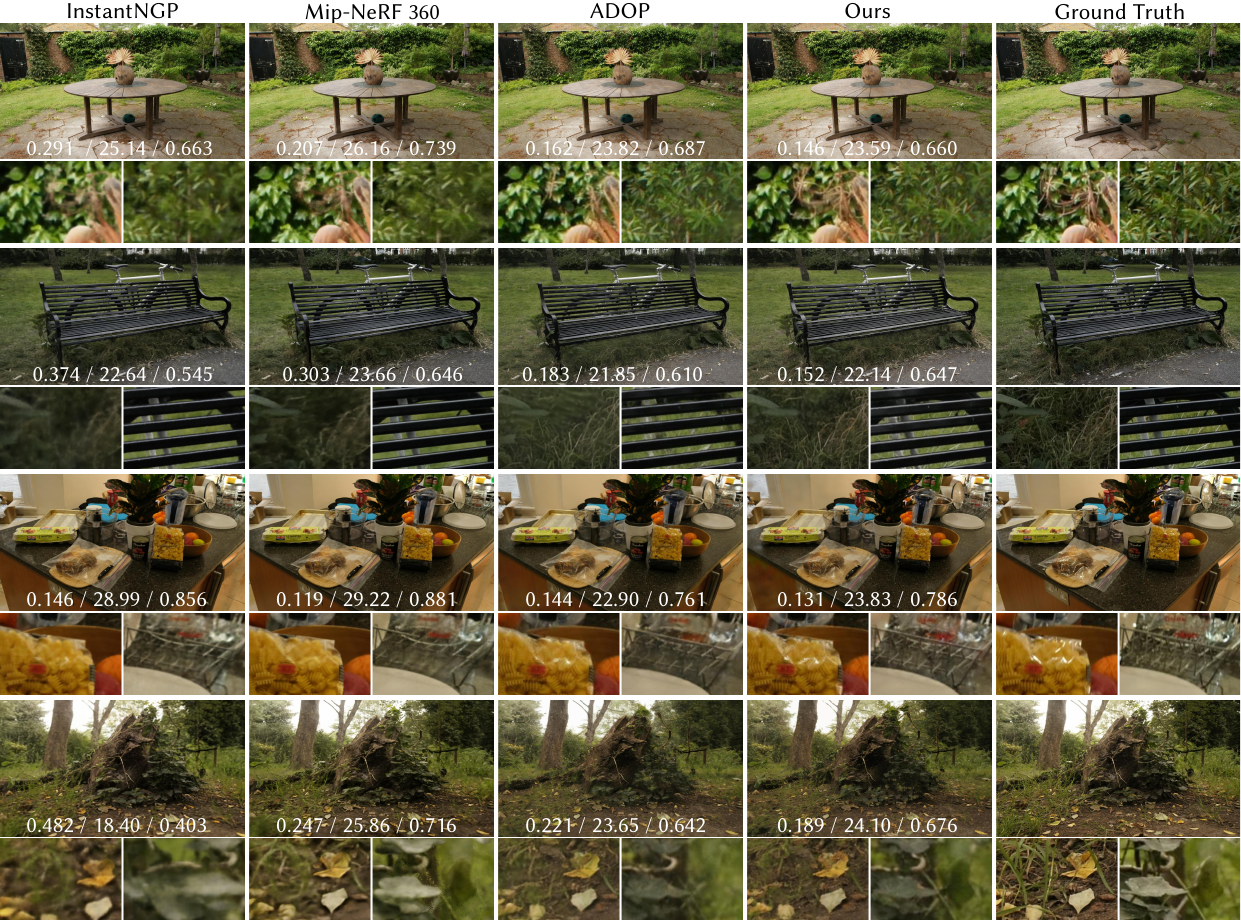}
    \centering
	\caption{Comparison on the MipNeRF-360 scenes with the average LPIPS $\downarrow$ / PSNR$\uparrow$ / SSIM$\uparrow$ scores denoted.}
	\label{fig:big-table}
\end{figure*}

\end{document}

%% file: 00-abstract.tex
\begin{abstract}
In the last few years, deep neural networks opened the doors for big advances in novel view synthesis.
Many of these approaches are based on a (coarse) proxy geometry obtained by structure from motion algorithms.
Small deficiencies in this proxy can be fixed by neural rendering, but larger holes or missing parts, as they commonly appear for thin structures or for glossy regions, still lead to distracting artifacts and temporal instability.
In this paper, we present a novel neural-rendering-based approach to detect and fix such deficiencies.
As a proxy, we use a point cloud, which allows us to easily remove outlier geometry and to fill in missing geometry without complicated topological operations.
Keys to our approach are (i) a differentiable, blending point-based renderer that can blend out redundant points, as well as (ii) the concept of \textit{Visual Error Tomography} (VET), which allows us to lift 2D error maps to identify 3D-regions lacking geometry and to spawn novel points accordingly.
Furthermore, (iii) by adding points as nested environment maps, our approach allows us to generate high-quality renderings of the surroundings in the same pipeline.
In our results, we show that our approach can improve the quality of a point cloud obtained by structure from motion and thus increase novel view synthesis quality significantly.
In contrast to point growing techniques, the approach can also fix large-scale holes and missing thin structures effectively.
Rendering quality outperforms state-of-the-art methods and temporal stability is significantly improved, while rendering is possible at real-time frame rates.
\end{abstract}

%% file: 01-intro.tex
\section{Introduction}
\label{sec:intro}

Recent advances in neural rendering have shown remarkable results on various tasks.
Especially the field of novel view synthesis~\cite{tewari2022advances} has received great attention from researchers around the world. 
\citet{Fridovich2022} have pointed out that the basis of the success comes from a differentiable rendering pipeline.
Thus, researchers have implemented inverse rendering formulations for several scene representations ranging from coordinate networks over volumetric data structures to triangle meshes or point clouds as geometric proxies.

In practice, explicit representations such as triangle meshes or point clouds are often preferred due to their simplicity to use and the possibility to integrate them into existing workflows.
Such proxies can be generated using 3D reconstruction techniques on input images or video, or using active 3D scanners.
To achieve high render quality, a high-quality proxy is required.
However, this is not always available as different capturing modalities fail in different scenarios.
Multi-view stereo depth estimation cannot reliably estimate featureless, reflective, or transparent surfaces, and often misses small, fine structures.
Active scanners, such as LIDAR, can handle textureless surfaces, but produce artifcats on dark and glossy surfaces and generally exhibit lower resolution.
As a result, the proxy geometry often is either incomplete or contains outliers.

In this paper, we propose a method to rectify these issues using a pair of \textit{cleaning} and \textit{spawning} steps.
As proxy we use a point cloud, which makes it easy to modify, add, and remove geometry without having to adapt topology.
Key to our approach is a differentiable point renderer that considers and optimizes point transparencies.
Points that receive high transparency are considered as outliers and can be removed without impact on quality.
This \textit{cleaning} operation enables us to improve the proxy point cloud twofold:
first, outliers from the original proxy, as they appear for instance in reflective regions or at object edges, can be removed to clean up the proxy and improve render performance.
Second, it opens the door for an effective \textit{spawning} operation:
We can tentatively add points in likely undersampled regions and let the cleaning step discard points that do not contribute to better renderings.
Thus, the surviving added points improve the proxy with the quality of this point completion being dependent on the accuracy of the prediction for undersampled regions.
To identify these critical regions in 3D space, we introduce a method we call Visual Error Tomography (VET).
Based on the output of the neural renderer and the corresponding input images, we generate error images and project these back to 3D space using computed tomography.
With this step, we can reliably identify 3D regions, where the original 3D reconstruction generated too few points, or no points at all, like at glossy surfaces or thin structures.
As we show in our results, this step is much more effective than point growing approaches \cite{furukawa2010towards,schoenberger2016mvs}, which tend to generate many unnecessary points, require many iterations to fill holes, and fail for thin structures.
We also show that by spawning (and cleaning) points in a few layers of nested environment maps, it is possible to faithfully re-render the environment, which improves rendering quality considerably.

The teaser image (Fig.~\ref{fig:teaser}) shows the point cloud for a tree obtained from multi-view stereo reconstruction along with the corresponding rendering result on the left. 
In the center, we present the enhanced point cloud, which notably enhances the rendering quality with other methods failing to reconstruct the fine-grain structure of the leafs faithfully.
In summary, our key contributions are:
\begin{itemize}
    \item A fast differentiable neural point-based rasterizer, capable of efficiently handling and optimizing point transparency
    \item The concept of Visual Error Tomography, a novel technique enabling reliable identification of erroneous 3D regions
    \item A pipeline incorporating iterative steps of point spawning and cleaning, resulting in the generation of clean and complete proxies, ultimately leading to a significant improvement in rendering quality
    \item The faithful visual reconstruction of the environment using nested point-based environment maps
    \item An evaluation of these methods, showing improved render quality particularly in difficult regions
    \item An open source implementation of the method located at:  \url{https://github.com/lfranke/VET}
\end{itemize}

%% file: 02-related.tex
\begin{figure*}
	\centering
	\includegraphics[width=.99\linewidth]{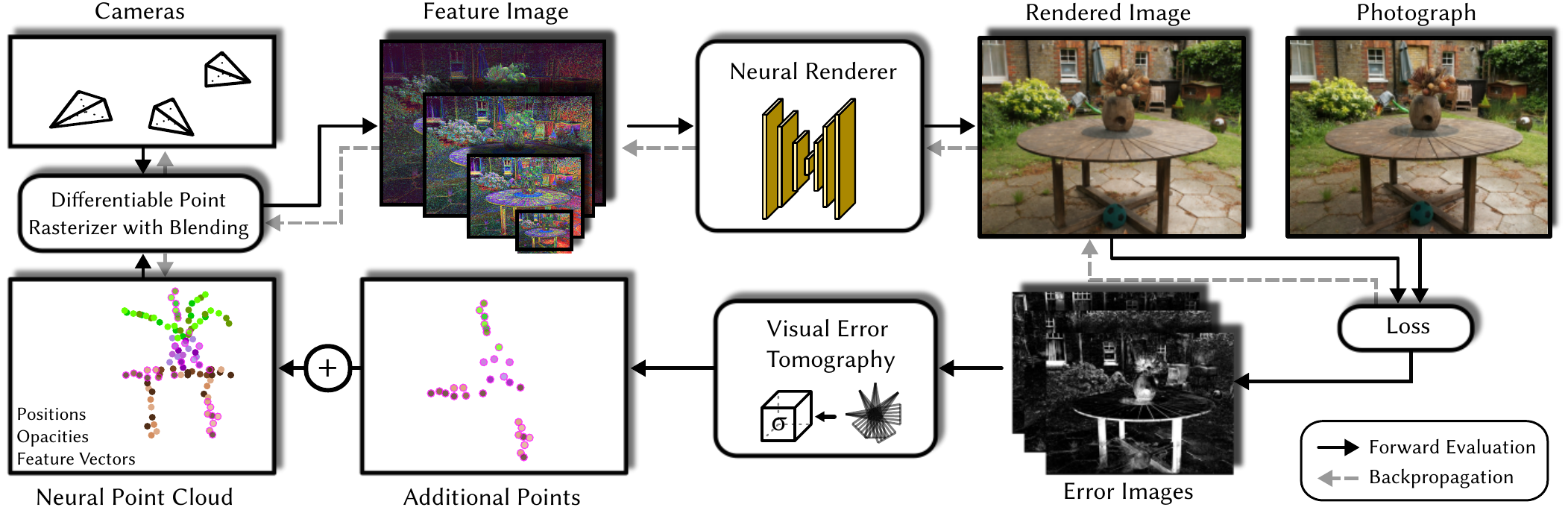}
    \centering
	\caption{
 Overview of our neural rendering pipeline with visual error tomography.
 Point descriptors, positions and opacities are optimized via backpropagation, which automatically removes outliers during training.
 After initial convergence, our VET is used to predict missing 3D points from the error images.
 Following, the optimization is resumed until the next convergence. 
 This process is repeated 1-3 times depending on scene complexity.
 }
	\label{fig:pipeline}
\end{figure*}

\section{Related Work}
\label{sec:related}

Novel view synthesis (NVS) is the task of generating new views of a scene given a set of input images. 
For example, one could capture a few photographs of an object and then use NVS to generate a video with an arbitrary camera trajectory. 
Many NVS implementations make use of image-based rendering techniques by directly warping the colors of several source images to a target view~\cite{debevec1998efficient,chaurasia2013depth}.
However, a geometrically correct warping requires an accurate proxy geometry, which is often not present on real datasets~\cite{shum2000review}.
Hence, traditional methods produce warping artifacts in rendered images especially around silhouettes.
These warping artifacts can be significantly reduced by replacing blending operation in the warping stage with deep neural networks~\cite{hedman2018deep,riegler2020free,riegler2021stable}.
If no proxy geometry is available, learning-based IBR approaches have been developed that create a multi-plane image representation~\cite{srinivasan2019pushing,tucker2020single,zhou2018stereo,mildenhall2019local} to directly estimate the warp field~\cite{ganin2016deepwarp,zhou2016view,flynn2016deepstereo}.
A disadvantage of IBR approaches is that all source images must be present in memory or quickly fetchable to render a novel view.

The desired solution to this problem is to generate a compact 3D representation of the scene in a preprocessing stage using only the source images as an input.
After that, novel views can be rendered efficiently without the need of all input images.
In the last couple of years, several 3D scene representations have been proposed, all with different advantages and disadvantages \cite{tewari2022advances}.
One of the most popular scene representations are coordinate networks, also known as NeRFs~\cite{mildenhall2021nerf}, since they are memory efficient, easy to use, and show excellent results on many datasets.
However, evaluating and training NeRF is time-consuming due to the need to evaluate a large neural network for each 3D sample.
This has lead to a series of subsequent works further improving quality~\cite{zhang2020nerf++,yu2021pixelnerf,martin2021nerf, barron2021mipnerf, barron2022mip, barron2023zip} or run-time~\cite{muller2022instant, reiser2021kilonerf, yu2021plenoctrees, turki2022mega}.

Next to coordinate networks, other scene representations such as sparse voxel grids~\cite{Fridovich2022}, octrees~\cite{ruckert2022neat}, tensor factorization~\cite{chen2022tensorf} and triangle meshes \cite{thies2019deferred} also show strong results on the NVS task.
Another promising scene representation are point clouds, since they can capture arbitrary fine geometric detail and are easy to use and modify.
Today it is possible to render billions of points in real time~\cite{schutz2022software} and simultaneously remove typical point-based rendering artifacts using deep neural networks~\cite{Pittaluga_2019_CVPR,meshry2019neural,song2020deep} and learned point descriptors~\cite{aliev2020neural,rakhimov2022npbg++, ruckert2022adop}.

A key observation to further improve the NVS quality is, that the point cloud and camera parameters used in these approaches can be optimized using differentiable rendering techniques.
In the context of point-based graphics, several approaches have been proposed for sphere rendering, Gaussian splatting, and point-sample rendering \cite{Lassner_2021_CVPR,wiles2020synsin,KPLD21,zhang2022differentiable,kopanas2022neural, kerbl20233d}.
A disadvantage of most of these approaches is that they do not have a reliable way to remove and spawn new points automatically.
This hinders rendering quality and leads to artefacts, if the initial point cloud is erroneous.
To overcome this problem, \citet{xu2022point} have introduced a point growing strategy to insert new points close to existing ones. 
Their approach demonstrates improved rendering quality on real datasets. 
Similarly, \citet{kerbl20233d} split large Gaussians in their pipeline to introduce new points.
Their point growing strategies however may not effectively handle regions where few existing points are available for reference, as well as issues with growth criteria selections~\cite{poux2022automatic} can be present.
To this end, Zuo and Deng~\shortcite{zuo2022view} propose an improved spawning technique, which identifies pixels with large rendering error and spawns new points randomly along the ray passing these pixels.
While this approach allows for reconstruction of missing objects, it is worth mentioning that only a limited number of points are placed on the object's surface. 
As a consequence, the reconstruction process can require thousands of iterations to complete and is prone to overfitting if points are spawned to close too the cameras.

Core of our method is \textit{visual error tomography} (VET), which is based on \textit{computed tomography} (CT). CT reconstructs 3D volumetric densities from a series of volumetric projections as obtained by x-ray imaging \cite{buzug2008computed}.
In our pipeline, we use Neural Adaptive Tomography (NeAT) \cite{ruckert2022neat} for CT reconstruction, because this approach can generate sharp, detailed reconstructions, and is able to model a non-linear intensity response of ray integrals, which is required for our non-physical reconstruction task.

%% file: 03-main.tex
\section{Overview}

Fig.~\ref{fig:pipeline} provides an overview of our pipeline. 
We start with a set of photographs, their corresponding poses, and an initial point cloud representing the scene.
Each point in the cloud is assigned a four-dimensional feature vector and an alpha value denoting opacity.
A differentiable point rasterizer renders the point cloud to feature images in multiple resolutions, which are then passed through a neural rendering U-Net to generate the final renderings \cite{aliev2020neural}.
The render loss is used to update point features and opacities via backpropagation. 
Depending on the initial quality of the reconstruction, optimization is performed on point positions and camera poses in this stage \cite{kopanas2022neural,ruckert2022adop}.
Points with low opacity are identified as outliers and automatically removed during the optimization.
Further details regarding the differentiable renderer and the cleaning procedure are provided in Sec.~\ref{sec:method-blend}.
Remaining image errors indicate missing geometry in the proxy.
To identify these undersampled regions, we apply our novel VET and spawn additional points in the found critical regions.
In Sec.~\ref{sec:spawn}, we provide details about VET and point spawning.
The entire process is repeated 1 to 3 times until convergence, dependent on scene complexity. 
As a result, we obtain a high-quality point cloud of the scene, which can be rendered by a real-time neural rendering system in photo-realistic quality.

\section{Point Cloud Optimization and Cleaning}
\label{sec:method-blend}

As initial point cloud we use the result generated by state-of-the-art 3D reconstruction pipelines such as COLMAP \cite{schoenberger2016mvs}.
In our examples, we use the densified point cloud, but as we show in the evaluation, starting with the sparse point cloud (similar to \citet{kerbl20233d}) and relying on our pipeline to fill in missing points still produces reasonable results.

To better represent the environment, previous methods usually use environment maps~\cite{zhang2020nerf++,ruckert2022adop}, recently in the form of multi-sphere images~\cite{Fridovich2022}.
In the same spirit, we spawn points on four layers of nested spherical environment maps using spherical Fibonacci sampling~\cite{keinert2015spherical,fink2019lumipath} of 0.5M points.
These points are optimized and cleaned identically to the original scene points.
In particular with the opacity bias for unseen points (Sec.~\ref{sec:conf_penalty}) this results in sparse and efficient environment maps and high-quality reconstructions of the surrounding.

\begin{figure*}
	\centering
    \includegraphics[width=.99\linewidth]{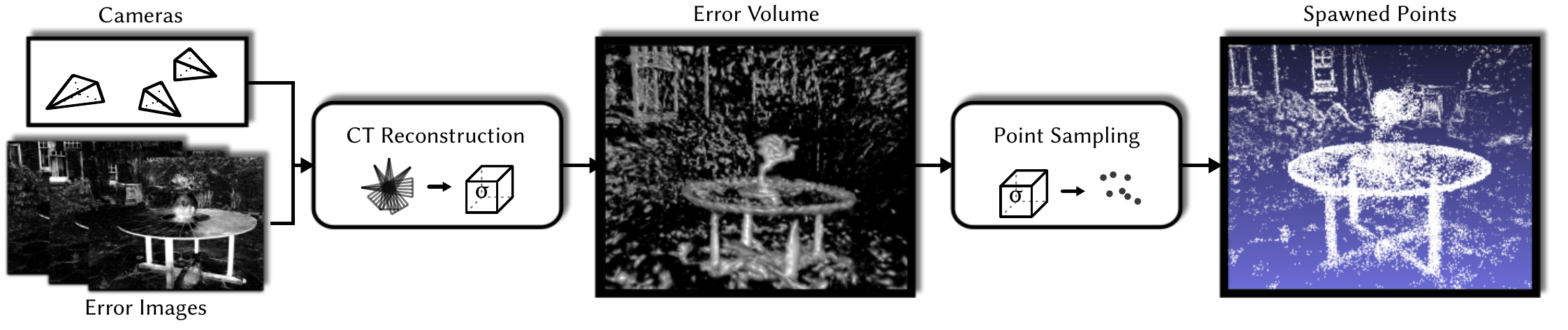}
    \centering
	\caption{Visual Error Tomography: starting from error images \textit{(left)}, we apply computed tomography to reconstruct a 3D error volume grid \textit{(center)}. Within each voxels, random points are spawned, where the number of spawned points depends on the voxel's error value \textit{(right)}.}
	\label{fig:vet_fig}
\end{figure*}
\subsection{Differentiable Point Rasterizer with Blending}

Our differentiable point-based rasterization approach is based on the work of \citet{ruckert2022adop}.
For detailed information on the forward and backward pass we refer to the original paper.
Our contribution to this pipeline is the use of front-to-back alpha blending instead of the fuzzy depth-test.
To that end, we first render the points to per-pixel linked lists, sort these lists by each point's depth, and finally blend the colors front to back by the respective alpha value. 
The final color $C_{u,v}$ with alpha blending is defined as
\begin{equation}
    C_{u,v} = \sum_{m=1}^{|\lambda|} T_m \cdot \alpha_m \cdot c_m 
      \label{eq:blend}
\end{equation} 
where $\lambda$ is the sorted per-pixel linked list, $\alpha_m$ the alpha value, and $c_m$ the respective color of point $m$.
The transmission variable $T_m$ describes how much light passes through all previous points in the list and therefore ensures that points at the end of the list only have a small contribution to the final color: $T_m =\prod_{i=1}^{m-1} (1-\alpha_i)$.
Since Eq.~\eqref{eq:blend} is differentiable, we can integrate this blending approach into the rasterization pipeline of \citet{ruckert2022adop} to optimize the alpha value of each point.
For a better optimization, alpha is stored in the range $(-\infty, \infty)$ and only converted to $[0, 1]$ during blending using the sigmoid function.

The rendering is performed four times with progessively lower resolutions and these are fed into a multi-scalar U-net with gated convolutions followed by a neural camera sensor module, as described by \citet{aliev2020neural} and \citet{ruckert2022adop}.
The multi-pixel rendering is implemented efficiently with a custom CUDA per-pixel counting and collection pass~\cite{selgrad2015filtering} followed by fast GPU bitonic sorting~\cite{batcher1968sorting,franke2018multi}.
For the backwards pass, only the sorted per-pixel lists have to be stored.

\subsection{Point Cleaning}

The previous optimization adapts the opacity of each point.
It is reasonable to assume that outlier points will receive low or zero opacity.
We thus determine outliers by introducing a threshold on the alpha value.
We found that during optimization, a threshold of $0.3$ works well on all tested scenes and gets rid of most outliers, while leaving room for variance during training.
For faster training, we remove points below this threshold every 50 epochs.

While previous neural point rendering approaches show some capability for disregarding outliers, mainly through strong masking abilities of gated convolutions, using this explicit optimization of a point confidence proves very powerful in removing outliers, which are either part of the initial reconstruction, environment spheres, or have been added by our spawning step.

\subsection{Point Opacity Bias}
\label{sec:conf_penalty}

In our point cloud optimization process, we add $\epsilon = 10^{-7}$ to the opacity gradient of each point.
This introduces a small bias to the opacity and encourages the system to remove all dispensable points, as unused points will be pushed towards a small opacity value. 
These points are often environment map points, far away outliers or inside structures.
This bias proves very powerful in reducing point cloud size without impacting rendering quality.

\subsection{Transition to Opaque Point Rendering}
\label{sec:transition}

Compared to direct one pixel point rendering with depth testing \cite{schutz2022software}, the alpha-based point blending required for outlier removal is slower to evaluate.
Therefore, if a real-time exploration of the scene is desired, we can smoothly transition our system from alpha blending to opaque depth testing during the optimization stage.
To this end, we adjust the sigmoid function, which encodes the raw alpha values~$\alpha_\text{raw}$, to become more steep over time:
\begin{equation}
\alpha = \text{sigmoid}\left( (10 + f \cdot t) \cdot \alpha_{\text{raw}}\right),
\label{eq:sigmoid}
\end{equation} 
where $f$ is a user defined parameter and $t$ the current optimization step. 
After the last iteration, the adjusted sigmoid is almost identical to the step function with each point having a binary opacity. 
For the final rendering, we then use only the opaque points without alpha blending, achieving high-performance real-time rendering of point clouds with 100 millions of points.

\section{Point Spawning via VET}
\label{sec:spawn}

The core of our point spawning step is visual error tomography (VET, Sec.~\ref{sec:method-ct}), which allows us to identify incomplete regions of the scene (see Fig.~\ref{fig:vet_fig}),
    where points then are spawned (Sec.~\ref{sec:spawning}).
We apply spawning multiple times when the neural rendering has converged sufficiently, usually every 200 epochs.

\subsection{Visual Error Tomography (VET)}
\label{sec:method-ct}

The idea of VET is to compute the per-pixel rendering loss of each input image and process these images with a CT reconstruction module.
The CT system creates a volumetric model that captures how much each point in space contributes to the visual error.
If the CT output for a given 3D point is small, most of the views must have a small visual error at its projected pixel. 
If the CT output is large, the visual error at the project pixel coordinates must also be large for multiple views.

Visual errors in neural point rendering often stem from missing points, local minima in descriptor optimization, or strong view dependency.
For all three, spawning primitives at that location is a way to improve results. For the first two cases, additional points fill the incomplete geometry or cause descriptor optimization to take additional training impulses.
For strong view dependency, adding points allows the blended descriptors to create a multi-layer surface representation that can approximate view-dependent effects.
While tackling this is not the explicit contribution of our pipeline (unlike e.g.~\citet{kopanas2022neural}), it does not hinder our VET's purpose.

To compute the error volume, we use the state-of-the-art CT reconstruction approach Neural Adaptive Tomography (NeAT)~\cite{ruckert2022neat}.
The major advantage of NeAT is its ability to model the non-linear intensity response of ray integrals, which is an essential property for our case because visual error maps inherently do not follow Beer-Lambert's law of photon attenuation.
Furthermore, the neural regularizer of NeAT prefers high density regions with sharp edges instead of large smooth areas.
This has provided better results for our method and produces sharp and clean error volumes, shown for example in Fig.~\ref{fig:vet_fig}.

For the input images, we smooth the error images $I$ with
\begin{equation}
    I'(x,y) = \text{clamp}(I(x,y)*(1+l) - l,0,1) 
        \label{eq:clamp_img}
\end{equation}  
to focus on high-error regions and ignoring small noise. We use $l=0.3$ for our error images. 

We observe that VET works well over the whole scene, including inside-out capturing scheme and backgrounds, but we limit the volume extent to keep memory costs low, as otherwise we would need to increase the output grid size (see next section).
For this bounding box we use 95\% of the inner points of the initial point cloud, as COLMAP or LIDAR commonly have extreme outliers, which are filtered out with this threshold.

\subsection{Error Volume Point Spawning}
\label{sec:spawning}
The output of VET is a $512^3$ voxel grid containing estimated error values.
After normalizing this grid to the range $[0,1]$, 
    we spawn $n_{(x,y,z)} = \lfloor e_i(x,y,z) \cdot p_\text{max} \rfloor$ new points, where $e_i$ is the normalized error
    value and $p_\text{max}$ the maximum number per voxel.
We use $p_\text{max}=10$, which typically leads to 10k - 2000k newly inserted points per VET step, but this is highly dependent on the error volume.
The positions of the $n$ added points are randomly distributed in the cell.
It is important to note that adding too many or inaccurately placed points is not an issue due to the robust outlier removal strategy described in the previous chapter.

Theoretically, this process converges and progressively fewer points are spawned each time, as less and less visual error is represented in the volume.
In practice, we spawn points 1-3 times during training, after which no further improvement can be observed.
Usually, our spawning and cleaning steps result in point clouds with a similar amount of points as the initial reconstruction, and thus have no significant impact on memory consumption and render times.

%% file: 04-eval.tex
\section{Evaluation}
\label{sec:evaluation}

\subsection{Datasets}
We have tested our approach on multiple scenes from  different datasets:
The Mip-NeRF 360 dataset~\cite{barron2022mip} has dense point clouds of 5M to 10M points and image resolutions of around $2500\times1600$ (half the original capturing).
In the Tanks\&Temples dataset~\cite{Knapitsch2017} images were captured with a high-quality RGB camera in full-HD and point clouds of 5M to 12M points.
The Sydney \textsc{Opera} House scene from~\citet{lu2023large} has 2.4M points and an image resolution of $1280\times676$.
The \textsc{Cherry Tree} was captured by us with a point cloud of 2M points and images of $1500\times1000$.
All initial reconstructions were done using the COLMAP multi-view stereo pipeline~\cite{schonberger2016structure}.
Additionally, the Redwood~\cite{Choi2016} dataset is used, which is captured with an RGB-D camera, with small image and point cloud resolutions.
For all experiments and methods, unless otherwise noted, 5\% of the images (every 20th) were left out of the training and used for evaluation. 

\subsection{Comparison to Related Work}
\label{sec:comprelated}

\begin{figure*}
	\centering
 \includegraphics[width=.99\linewidth]{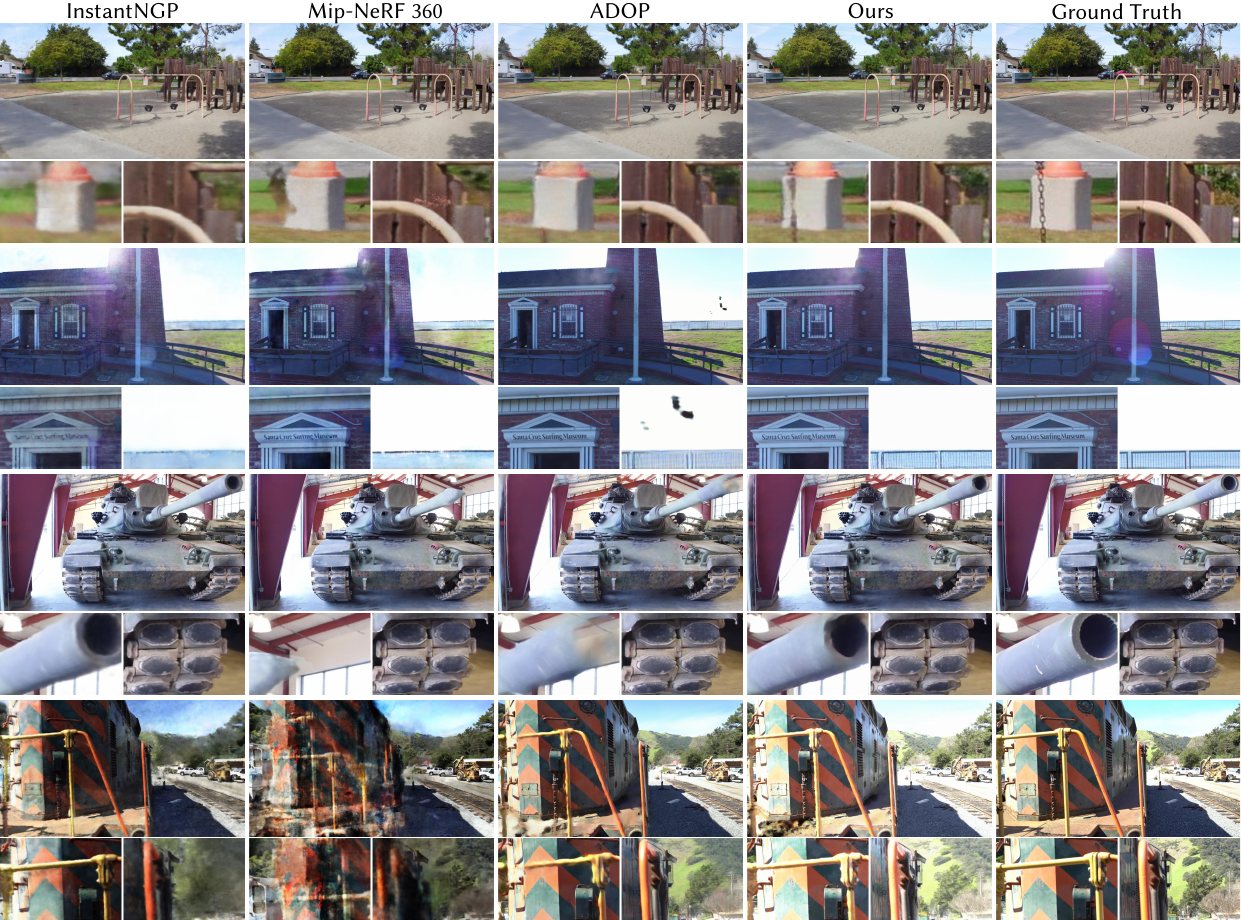}
    \centering
	\caption{Comparison to prior work on the Tanks\&Temples dataset. Our method is able to reconstruct fine details, remove outliers and improve overall sharpness.}
	\label{fig:big_tt}
\end{figure*}

To evaluate the neural rendering performance of our pipeline we have measured the inference quality on several scenes.
We compare against state of the art approaches: Three NeRF-based with InstantNGP~\cite{muller2022instant} (with their larger configuration), Mip-NeRF 360~\cite{barron2022mip} and Nerf++~\cite{zhang2020nerf++}, a triangle-based approach in Stable View Synthesis~\cite{riegler2021stable} and a point-based approach with ADOP~\cite{ruckert2022adop}.
Fig.~\ref{fig:big_tt} shows a visual comparison on the Tanks\&Temples dataset, with the quantitative evaluation presented in Tab.~\ref{tab:eval_tt}.
On all scenes, our method outperforms the compared approaches due to our VET-based point cloud completion.
Especially noteworthy is our approach's ability to reconstruct fine detail as well as overall sharpness.
In Fig.~\ref{fig:big-table}, we showcase four scenes from the Mip-NeRF 360 dataset, observing the same characteristics.
Here however, our PSNR scores tend to be lower than Mip-NeRF 360 as slight errors in calibrations and color shifts by the CNN impact this metric proportionally higher while not impacting perceived visual quality~\cite{zuo2022view}.
Our rendering times however are significantly faster than theirs, as seen in Tab.~\ref{tab:times}.

Video results can be found at: \url{https://youtu.be/adH6GyqC4Jk}

\begin{table}[]
\caption{\label{tab:times}%
Training and render times on the \textsc{Garden} and \textsc{Playground} scenes. The VET reconstruction takes about 15 minutes and is included in our training time. Inference measured on an RTX4090, training on an A100. }
\footnotesize\centering%
\begin{tabular}{l|r|r| r} 
\textit{Method} &  Training & Render \textsc{(Garden)} & Render \textsc{(Playground)} \\ \hline
InstantNGP (8 spp)  & 0.25 h & 1046 ms  & 1372 ms \\
Mip-NeRF 360 & 36 h & 38390 ms & 18240 ms \\
ADOP & 8 h & 30 ms & 15 ms  \\
Ours (Blend) & 16 h &  33 ms &  23 ms \\
Ours (Opaque) & 16 h &  29 ms & 13 ms \\
\end{tabular}%
\vspace{-1mm}
\end{table}

\begin{table*}[]
\centering
\caption{Results on the Tanks\&Temples dataset. See also Fig.~\ref{fig:big_tt} for a visual comparison.}%
\footnotesize\begin{tabular}{@{}l|ccc|ccc|ccc|ccc@{}}%
\toprule
                      & \multicolumn{3}{c|}{\textsc{Train}}                                  & \multicolumn{3}{c|}{\textsc{Playground}}                               & \multicolumn{3}{c|}{\textsc{M60}}                                      & \multicolumn{3}{c}{\textsc{Lighthouse}}                            \\
\textit{Method}        &  LPIPS $\downarrow$  & PSNR $\uparrow$  & SSIM $\uparrow$ &  LPIPS $\downarrow$  & PSNR $\uparrow$   & SSIM $\uparrow$ &  LPIPS $\downarrow$  & PSNR $\uparrow$ & SSIM $\uparrow$   &  LPIPS $\downarrow$  & PSNR $\uparrow$ & SSIM $\uparrow$\\ \midrule
NeRF++          & 0.434  & 18.05 & 0.579 & 0.441  & 22.25 & 0.613 & 0.360 & 23.06 & 0.728  & 0.386  & 20.08 & 0.662\\
SVS (half res.) & 0.267  & 17.44 & 0.683 & 0.291  & 22.12 & 0.704 & 0.197 & 23.74 & 0.831  & 0.264  & 17.14 & 0.722\\
InstantNGP      & 0.335  & 20.46 & 0.644 & 0.418  & 18.69 & 0.518 & 0.203 & 24.97 & 0.797  & 0.285  & 22.98 & 0.728\\
Mip-NeRF 360    & 0.355  & 18.37 & 0.624 & 0.271  & 24.86 & 0.736 & 0.189 & 24.68 & 0.838  & 0.235  & 21.35 & 0.750\\
ADOP            & 0.131  & 21.44 & 0.723 & 0.126  & 24.85 & 0.719 & 0.103 & 24.91 & 0.811  & 0.119  & 22.27 & 0.755\\\midrule
Ours & \textbf{0.123}  & \textbf{21.67} & \textbf{0.765} & \textbf{0.124}   & \textbf{25.49} & \textbf{0.768} & \textbf{0.080}  & \textbf{27.34} & \textbf{0.875} & \textbf{0.094}  & \textbf{24.19} & \textbf{0.808} \\\bottomrule
\end{tabular}
\label{tab:eval_tt}%
\end{table*}

\subsection{Ablation Studies}

\subsubsection{Evaluation of the Error Metric used in VET}
The VET module takes visual error maps as input to compute the error volume.
In Tab.~\ref{abl:tomography-loss}, we show rendering quality with different VET error metrics on the tree scene.
Overall, using SSIM provides the best results as it guides more points to be spawned in thin locations.

\begin{table}[]
\parbox[t][][t]{.42\linewidth}{%
\caption{Rendering error when different visual error metrics are used for the VET reconstruction on the \textsc{Cherry Tree} scene. L2 error is processed with $l=0.01$.}%
\label{abl:tomography-loss}%
\centering\footnotesize%
\begin{tabular}[t]{l|ccc}
\textit{VET-loss} & LPIPS  $\downarrow$ & PSNR $\uparrow$  \\ \hline 
use L1        & 0.177               & 21.57           \\
use L2        & 0.187               & 21.19          \\
use SSIM      & 0.162               &  21.88         \\   
\end{tabular}
}\hfill
\parbox[t][][t]{.54\linewidth}{%
\caption{Sigmoid narrowing factor $f$ used for transitioning to depth-test based rendering on the Redwood \textsc{Chair} scene. After 400 epochs, the switch was initiated, final result after 600 epochs.}%
\label{abl:narrowing-fac}%
\footnotesize\centering%
\begin{tabular}[t]{l|cc}
$f$         & LPIPS  $\downarrow$ & PSNR $\uparrow$  \\ \hline  
$10^{0}$    & 0.200  & 18.85             \\ 
$10^{-1}$   & 0.197  & 18.90             \\ 
$10^{-2}$   & 0.186  & 19.66            \\ 
$10^{-3}$   & 0.191  & 19.44            \\  
\end{tabular}
}%
\end{table}

\subsubsection{Narrowing Factor}
In Sec.~\ref{sec:transition}, we have introduced a sigmoid narrowing factor to smoothly transition the alpha renderer to traditional depth testing.
Tab.~\ref{abl:narrowing-fac} shows the experiment with different values of $f$, with large values impacting quality, as changes are too abrupt during training.
We therefore recommend $f=0.01$.

\begin{figure}
\centering%
\includegraphics[width=.8\linewidth]{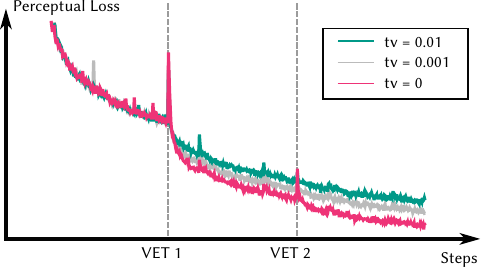}%
\caption{Loss graph for running VET with different values of the TV regularizer.}%
\label{fig:tv-reg}%
\end{figure}

\subsubsection{TV Regularization in VET}
The neural CT reconstruction approach uses a total variation (TV) regularizer to smooth the output volume.
We have tested our pipeline with different TV values and found that with a smaller value, more outlier points are spawned.
However, since these points are cleaned reliably, the difference in the final result is only marginal (see Fig.~\ref{fig:tv-reg}).
For the remaining experiments we use $tv=0.001$ as it is a good compromise between capturing small details and not spawning too many outliers.

\begin{figure}
\centering%
\includegraphics[width=.92\linewidth]{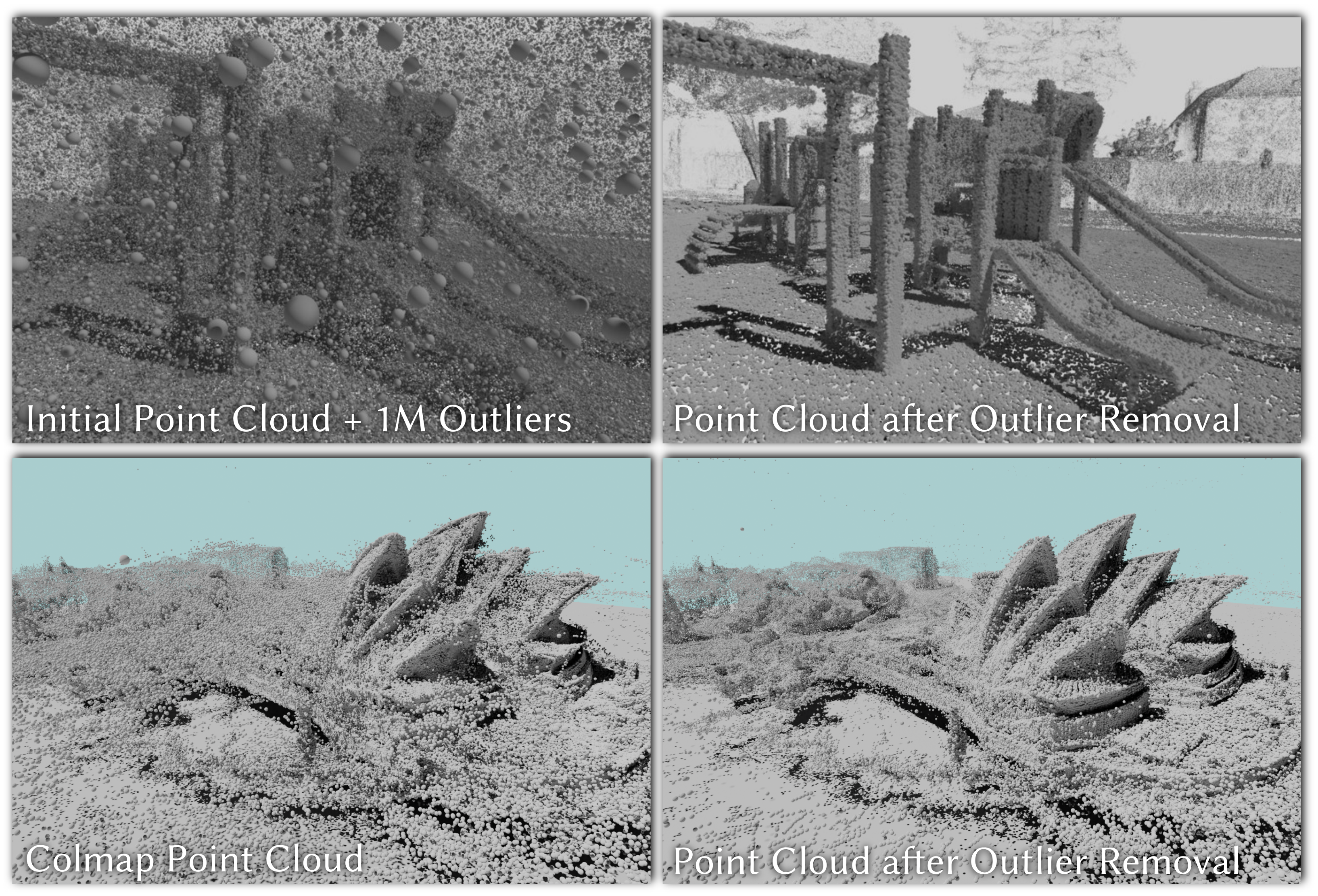}%
\caption{Ablation study on point cleaning. \textit{(Top)} 1M random points have been added to the \textsc{Playground} scene \textit{(left)} and progressively cleaned \textit{(right)}. On the \textsc{Opera} scene \textit{(bottom)}, COLMAP has placed points originating from the water incorrectly on land \textit{(left)}, which our pipeline can fix \textit{(right)}. }%
\label{fig:outlier_removal}%
\end{figure}

\subsubsection{Point Cleaning}
To test the robustness of the outlier cleaning strategy, we have ran our pipeline on the \textsc{Playground} scene, where we have added one million additional random points around the playground items.
As shown in Fig.~\ref{fig:outlier_removal} \textit{(top)} on the right, the final point cloud contains almost no visible outliers. 
On the \textsc{Opera} scene (Fig.~\ref{fig:outlier_removal} \textit{bottom}), this robustness provides great cleanup in water areas, where outliers are placed over the water surface through COLMAP's reconstruction.

\subsubsection{Point Opacity Bias}
In Sec.~\ref{sec:conf_penalty}, we introduced the opacity bias for the points.
As seen in Tab.~\ref{tab:point-gradient-penalty}, as long as the bias is not chosen extremely strong, low rendering loss is maintained while the amount of points is efficiently lowered. 
We choose $10^{-7}$ for our experiments, however we note that $10^{-6}$ reduces the point cloud efficiently while still providing sensible, if slightly worse results.

\begin{table}[]
\parbox[t][][t]{.48\linewidth}{%
\caption{\label{tab:point-gradient-penalty}%
Point opacity bias ablation study on the \textsc{Garden} scene (7.8M points). 
Higher bias aggressively removing points while still maintaining low rendering loss.
}
\footnotesize\centering%
\begin{tabular}[]{l|ccc} 
\textit{Bias}&Points&LPIPS $\downarrow$&PSNR $\uparrow$\\ \hline
$10^{-5}$& 0.2M  & 0.406  & 20.73  \\
$10^{-6}$& 4.3M  & 0.186  & 23.60   \\
$10^{-7}$& 7.3M  & 0.178  & 23.70    \\
$10^{-8}$& 8.0M  & 0.177  & 23.71    \\
$10^{-9}$& 8.9M  & 0.178  & 23.67   \\
$10^{-10}$& 9.6M  & 0.178  & 23.67  \\ 
\end{tabular}
}\hfill
\parbox[t][][t]{.48\linewidth}{%
\caption{\label{tab:point-cleaning-threshold}%
Point cleaning threshold on the \textsc{Garden} scene (7.8M points).
Point clouds can be cleaned aggressively before impacting rendered results.
}%
\footnotesize\centering%
\begin{tabular}[]{l|ccc} 
$T$ &Points& LPIPS $\downarrow$ & PSNR $\uparrow$ \\ \hline
0.10 & 10.3M  & 0.176  & 23.74  \\
0.20 & 9.1M  & 0.177  & 23.69   \\
0.30 & 7.3M  & 0.178  & 23.70    \\
0.50 & 6.6M  & 0.179  & 23.62    \\
0.70 & 5.3M  & 0.183  & 23.67   \\
0.90 & 2.6M  & 0.200  & 23.44  \\ 
0.99 & 0.1M  & 0.446  & 19.08  \\ 
\end{tabular}
}
\end{table}

\begin{table}[]
\caption{\label{tab:pmax_abl}%
Amount of points spawned compared with rendering losses.}%
\footnotesize\centering
\begin{tabular}{l|ccc|ccc} 
   & \multicolumn{3}{c|}{\textsc{Garden} (Init. 7.8M Points)} & \multicolumn{3}{c}{\textsc{Cherry Tree} (Init. 2M Points)} \\
$p_{max}$ &  \# Points  & LPIPS $\downarrow$ & PSNR $\uparrow$ &  \# Points  & LPIPS $\downarrow$ & PSNR $\uparrow$ \\ \hline
2 & 7.0M  & 0.181   & 23.60  & 1.7M & 0.295 & 18.74\\
5 & 7.2M  & 0.180   & 23.69 & 1.9M& 0.280 & 19.39\\
10 & 7.3M  & 0.178  & 23.70 & 4.0M& 0.223 & 20.41\\
20 & 7.8M  & 0.177  & 23.71 & 17.3M& 0.182 & 21.33\\
30 & 8.5M  & 0.177  & 23.73 & 31.5M& 0.167 & 21.88\\ 
\end{tabular}
\end{table}

\subsubsection{Point Cleaning Confidence Threshold}
In every cleaning step, points with confidence below the threshold are removed.
As described in Tab.~\ref{tab:point-cleaning-threshold}, thresholds up to $0.9$ still provide reasonable results and reduce point cloud size significantly.
However, we use $0.3$ for our experiments as it causes point clouds to be similarly sized as the initial COLMAP reconstruction.

\subsubsection{Number of Spawned Points per Voxel}
As described in Sec.~\ref{sec:spawning} we spawn up to $p_{max}$ points per voxel in the error volume.
In Tab.~\ref{tab:pmax_abl}, the resulting point cloud and the rendering loss is evaluated.
We found that $p_{max}=10$ works well in most cases and keeps total point numbers similar to initial COLMAP reconstructions.
Thus we use this for all experiments.
We note however that for difficult cases where COLMAP fails considerably, a high $p_{max}$ can improve quality significantly (e.g. in the cherry tree scene, Fig.~\ref{fig:teaser}).

\subsubsection{Environment Map}

\begin{table}[]
\caption{\label{tab:environment}%
Environment map setup on the \textsc{Playground} scene (12.5M Points).
}%
\footnotesize\centering%
\begin{tabular}{l|ccc} 
\textit{Method} &  \# Points  & LPIPS $\downarrow$ & PSNR $\uparrow$ \\ \hline
No Environment Map & 10.0M  & 0.146  & 25.34  \\
1 Textured Sphere & 11.4M  & 0.176  & 24.99   \\
4 Layered Textured Spheres & 11.3M  & 0.181  & 25.00    \\
1 Point-based Sphere (init. 0.5M points) & 10.8M  & 0.151  & 24.59    \\
4 P.-b. Layered Spheres (init. 2.0M points) & 13.3M  & 0.124  & 25.49   \\ 
\end{tabular}
\end{table}
As described, we use four layered point-based spheres for our environment map. 
When compared with different common environment map strategies (see Tab.~\ref{tab:environment}), ours (four point-based spheres) performs best when compared to one point-based sphere or to texture-based environment maps with one~\cite{ruckert2022adop} or four layers~\cite{yu2021plenoctrees}.
In contrast to textured maps, our method tends to be less susceptible to overfitting on training view, which in turn improves VET as the error images are more meaningful.

\subsection{Point Spawning Accuracy}
\label{sec:spawningaccu}
\begin{figure}%
\footnotesize\centering%
\begin{tabular}{l|ccc|ccc}
   & \multicolumn{3}{c}{\textsc{Meetingroom}} & \multicolumn{3}{c}{\textsc{Barn}}\\
\textit{Method}            & Prec. $\uparrow$ & Rec. $\uparrow$ & \textbf{F-score $\uparrow$}  & Prec. $\uparrow$  & Rec. $\uparrow$ & \textbf{F-score $\uparrow$} \\ \hline
COLMAP      & 0.521 & 0.247 & \textbf{0.335}  & 0.633 & 0.553 & \textbf{0.591} \\
Ours        & 0.507 & 0.253 & \textbf{0.338}  & 0.631 & 0.555 & \textbf{0.591}  \\
Ours-full   & 0.512 & 0.283 & \textbf{0.365}  & 0.624 & 0.679 & \textbf{0.651}  \\
\end{tabular}
\includegraphics[width=\linewidth]{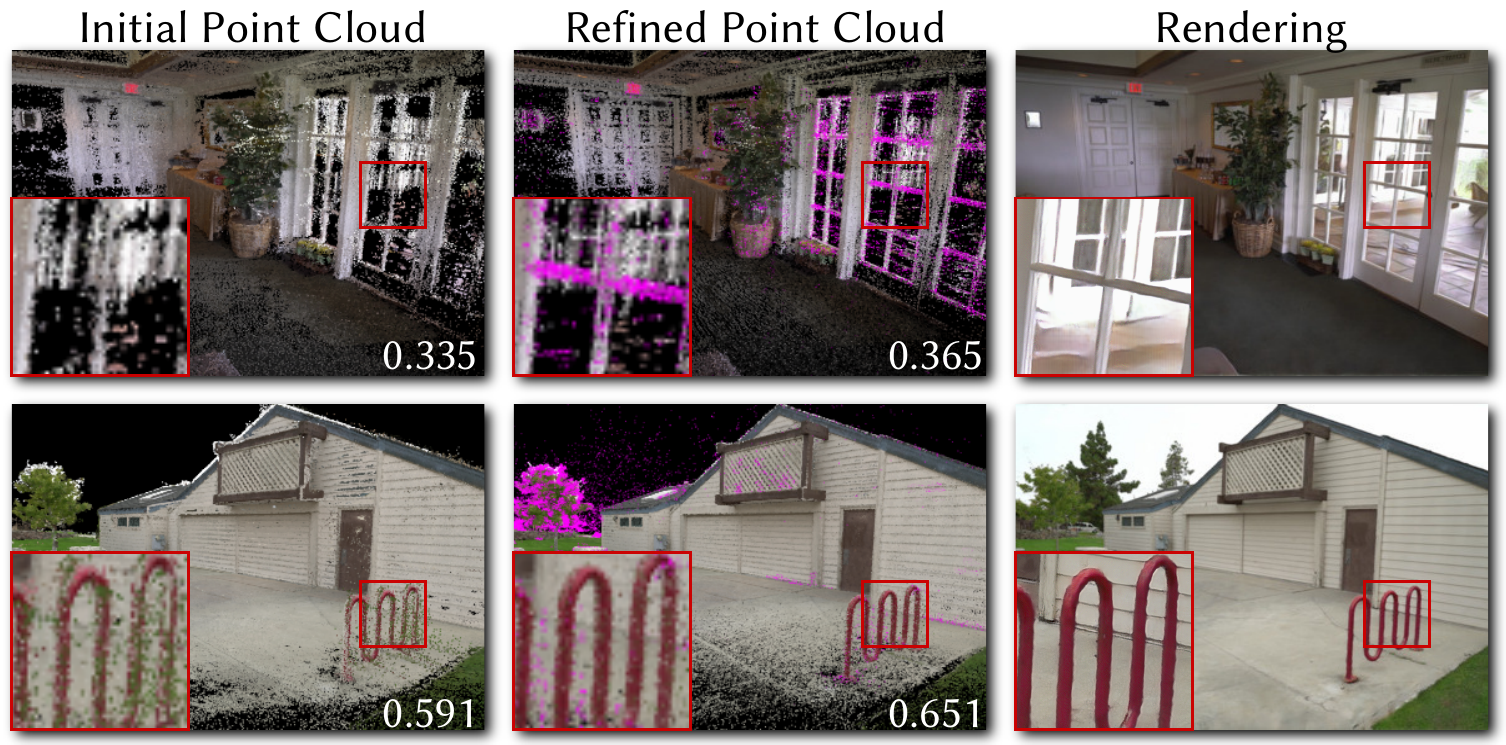}
\centering
\caption{Geometric accuracy, measured with the F-score metric~\cite{Knapitsch2017}. 
Our approach as described (\textit{Ours}) provides similar results in geometry accuracy as COLMAP, while our approach without cleaning the initial point cloud (\textit{Ours-full}) improves scores by about 10\%.
}
\label{fig:tt_fscore}
\end{figure}

While our method does not explicitly try to increase geometric accuracy of the input model (but tries to minimize visual loss for novel view synthesis), we evaluate this accuracy on two training scenes of the popular Tanks\&Temples benchmark with their \textit{F-score} based metric~\cite{Knapitsch2017}.
As seen in Fig.~\ref{fig:tt_fscore}, the point cloud is improved by almost 10\% compared to COLMAP if the initial dense reconstruction is not cleaned (\textit{ours-full}), thus only VET-spawned points are potentially removed.
Otherwise, with our normal pipeline, scores are very similar as geometrically accurate points are spawned, but initial correctly placed points might be cleaned if they are not important for visual results of our neural renderer.

\subsection{Sparse Colmap Reconstruction as Input}
\label{sec:sparse_points}
\begin{figure}
	\centering
    \includegraphics[width=.99\linewidth]{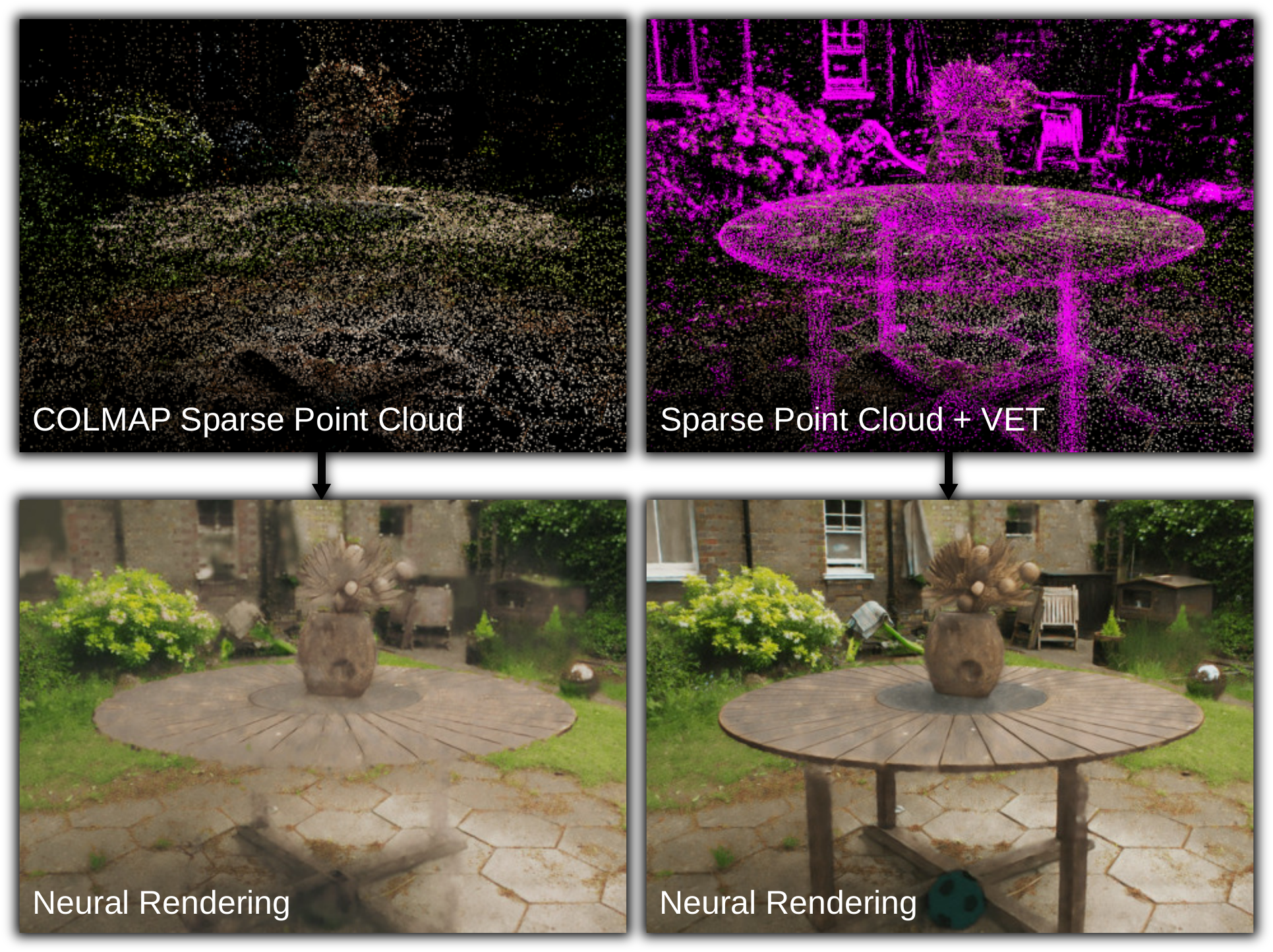}
    \centering
	\caption{Starting with COLMAPs sparse reconstruction in our pipeline.
 Neural rendering without VET \textit{(left)} converges to LPIPS values of $0.596$, with VET~\textit{(right)} this improves to $0.341$.
    }
	\label{fig:sparse_reco}
\end{figure}

We can also skip COLMAP's dense reconstruction and directly work with the sparse reconstruction (similar to Gaussian splatting's pipeline~\cite{kerbl20233d}), thus starting with a point cloud of only 130K points.
In this setup, we augment our pipeline with an additional spawning step after 75 epochs, as this low amount of points reaches convergence sooner.
As seen in Fig.~\ref{fig:sparse_reco}, the resulting point cloud and neural rendering provides overall good results. 

\begin{table}[]%
\caption{\label{tab:snp_pnerf}%
Comparison with Point-NeRF~\cite{xu2022point} and SNP~\cite{zuo2022view} on their cropped Tanks\&Temples dataset.}%
\footnotesize\centering%
\begin{tabular}{l|ccc|ccc} 
   & \multicolumn{3}{c|}{\textsc{Barn} (Fig.~\ref{fig:point_spawn})} & \multicolumn{3}{c}{Avg. Tanks\&Temples} \\
Method &  LPIPS $\downarrow$ & PSNR $\uparrow$ &  SSIM $\uparrow$  & LPIPS $\downarrow$ & PSNR $\uparrow$ &  SSIM $\uparrow$\\ \hline
Point-NeRF & 0.120  & 29.15   & 0.937  & 0.080 & 29.61 & 0.954\\
SNP & 0.109  & 29.80   & 0.915 & 0.079& 29.78 & 0.942\\
Ours & 0.024  & 29.66  & 0.939 & 0.026 & 29.54 & 0.950\\
\end{tabular}
\end{table}

\begin{figure}
    \includegraphics[width=\linewidth]{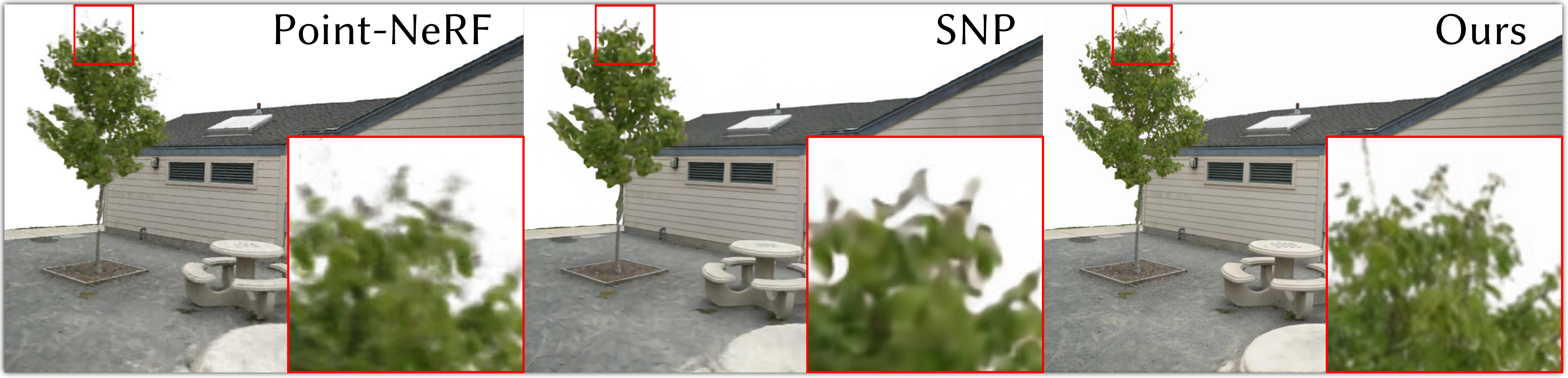}
    \centering
	\caption{Different point spawning novel synthesis methods. Both Point-NeRF~\cite{xu2022point} and View Synthesis with Sculpted Neural Points (SNP)~\cite{zuo2022view} use different point spawning strategies with growing and error projection. Both however fail to reconstruct fine details, as with the leaves of the tree.}
	\label{fig:point_spawn}
\end{figure}

\subsection{Comparison to Different Spawning Strategies}
\label{sec:reg_grow_exp}
\label{sec:spawningcomp}

\subsubsection{PointNeRF and SNP}
In Fig.~\ref{fig:point_spawn}, we compare to two other novel view synthesis methods using point cloud augmentations on their cropped version of the Tanks\&Temples \textsc{Barn} with PointNeRF's train/eval split.
PointNeRF~\cite{xu2022point} spawns points via growing and SNP~\cite{zuo2022view} samples error map regions continuously in space.
Both however fail to reconstruct small details such as the leaves.
This is also reflected in the quantitative measurements in Tab.~\ref{tab:snp_pnerf}, where PSNR and SSIM scores are similar, while we ourperform both in LPIPS scores.

\subsubsection{Point Growing in our Framework}
\begin{figure}
	\centering
    \includegraphics[width=\linewidth]{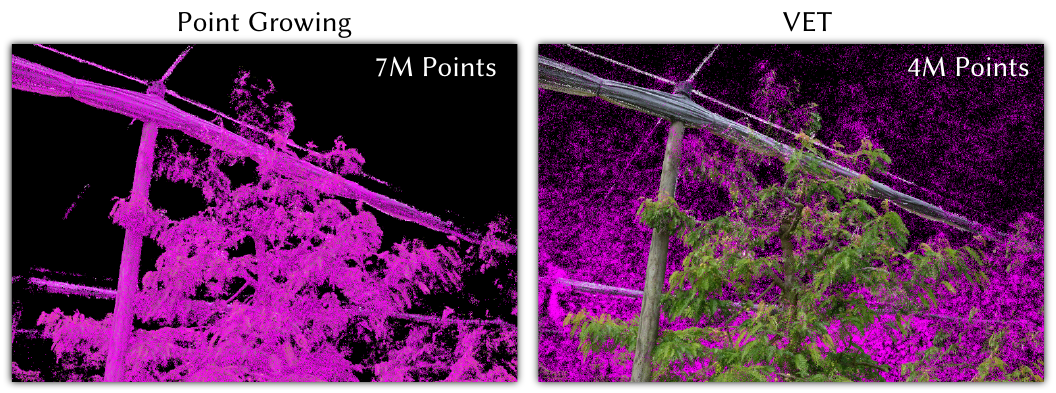}
    \centering%
	\caption{Point cloud of the \textsc{Cherry Tree} enhanced with a point growing strategy \textit{(left)} and our VET \textit{(right)}.}
	\label{fig:growing_tree}
\end{figure}
This result matches with our observations.
A point growing strategy implemented in our pipeline instead of VET has the same difficulties while increasing point cloud sizes significantly compared to VET, as seen in Fig.~\ref{fig:growing_tree}.
Furthermore unprojecting pixel errors and summing them into voxels (similar to SNP's strategy) causes point spawns being scattered over larger regions.
Solving this either requires thresholding which causes details to be missed or requires to use aggressive point cleaning impacting quality.
VET however avoids all these problems.

\subsection{View Dependent Effects}
\label{sec:view_dependancy}
\begin{figure}
	\centering
	\includegraphics[width=.99\linewidth]{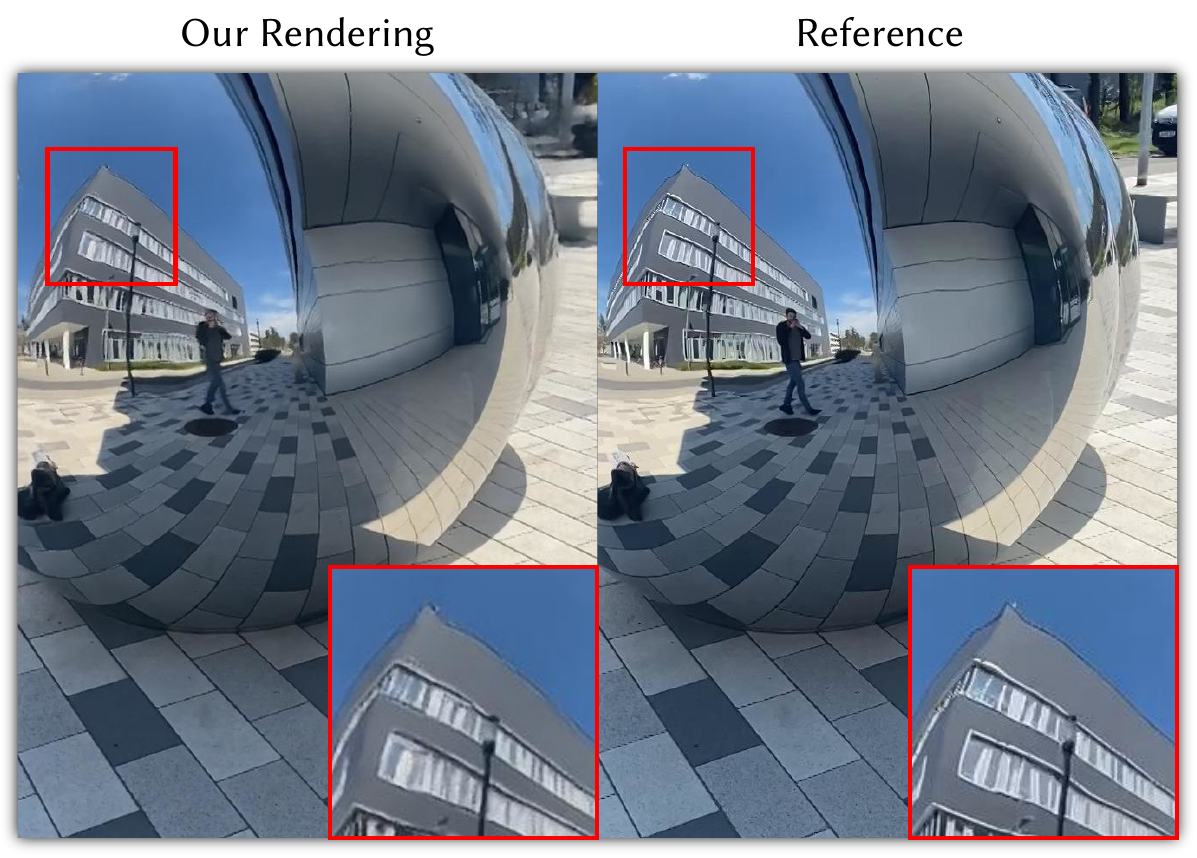}
    \centering
	\caption{\textsc{Reflecting Spheres}. Our approach can approximate reflections after point cloud cleaning.
    \label{fig:spheres}}%
\end{figure}
While not explicitly modeled in our approach, our blending renderer can handle view dependencies relatively well (as seen in Fig.~\ref{fig:spheres}) by optimizing reflections into multiple point layers.
Note however, that very good angular coverage during training is necessary and explicitly modeled approaches such as~\citet{kopanas2022neural} may be preferable in this use case.

%% file: 05-conc.tex
\section{Limitations}
\label{sec:limitations}

We have encountered a few limitations, inherited from the one-pixel point rendering \cite{aliev2020neural}, which is efficient but produces artifacts in rare cases.
When the camera gets too close to a surface, the neural rendering network is not able to close the holes between the points and thus the surface behind the wall becomes visible.

Furthermore, we have observed some temporal instabilities when the camera is moved through the scene slowly, which is a side effect of the discrete one-pixel point rendering because points can only discretely move from one pixel to the next.

\section{Conclusion}

To conclude this paper, we have proposed a point-based neural rendering pipeline that is able to render photo-realistic novel views and can simultaneously refine and complete the input point cloud.
Point spawning is implemented using our novel VET technique, which uses the visual error maps to identify and populate undersampled regions of the scene.
Outlier points are removed automatically by thresholding their opacity value, which is optimized during the training stage.
We have shown in several experiments that our pipeline produces high-quality point clouds, and outperforms current state-of-the-art neural rendering approaches in quality.
The source code is available at:
\begin{center}
    \url{https://github.com/lfranke/VET}
\end{center}